\documentclass{article}
\usepackage{ijcai21}

\usepackage{times}
\usepackage{soul}
\usepackage{url}
\usepackage[hidelinks]{hyperref}
\usepackage[utf8]{inputenc}
\usepackage[small]{caption}
\usepackage{graphicx}
\usepackage{subfigure}
\usepackage{amsmath}
\usepackage{amsthm}
\usepackage{amssymb}
\usepackage{booktabs}
\usepackage{algorithm}
\usepackage{algorithmic}

\usepackage{listings}
\usepackage{color}

\definecolor{dkgreen}{rgb}{0,0.6,0}
\definecolor{gray}{rgb}{0.5,0.5,0.5}
\definecolor{mauve}{rgb}{0.58,0,0.82}

\title{Augmenting Proposals by the Detector Itself}

\author{
Xiaopei Wan$^1$
\and
Zhenhua Guo$^2$\and
Chao He$^{1}$\and
Yujiu Yang$^1$\and
Fangbo Tao$^2$
\affiliations
$^1$Tsinghua University\\
$^2$Alibaba Group\\
\emails
\{wxp18, hec18\}@mails.tsinghua.edu.cn,
yang.yujiu@sz.tsinghua.edu.cn,
\{mianzhang.gzh, fangbo.tfb\}@alibaba-inc.com
}

\begin{document}

\maketitle

\begin{abstract}
  Lacking enough high quality proposals for RoI box head has impeded two-stage and multi-stage object detectors for
  a long time, and many previous works try to solve it via improving RPN's performance or manually generating proposals 
  from ground truth. However, these methods either need huge training and inference costs or bring little improvements. 
  In this paper, we design a novel training method named APDI, which means augmenting proposals by the detector itself 
  and can generate proposals with higher quality. Furthermore, APDI makes it possible to integrate IoU head into RoI 
  box head. And it does not add any hyper-parameter, which is beneficial for future research and downstream tasks. 
  Extensive experiments on COCO dataset show that our method brings at least 2.7 AP improvements on Faster R-CNN with 
  various backbones, and APDI can cooperate with advanced RPNs, such as GA-RPN and Cascade RPN, to obtain extra gains. 
  Furthermore, it brings significant improvements on Cascade R-CNN.
  \footnote{All source codes will be released after publication.}
\end{abstract}

\section{Introduction}

\begin{figure}
    \centering
    \includegraphics[width=0.99\linewidth]{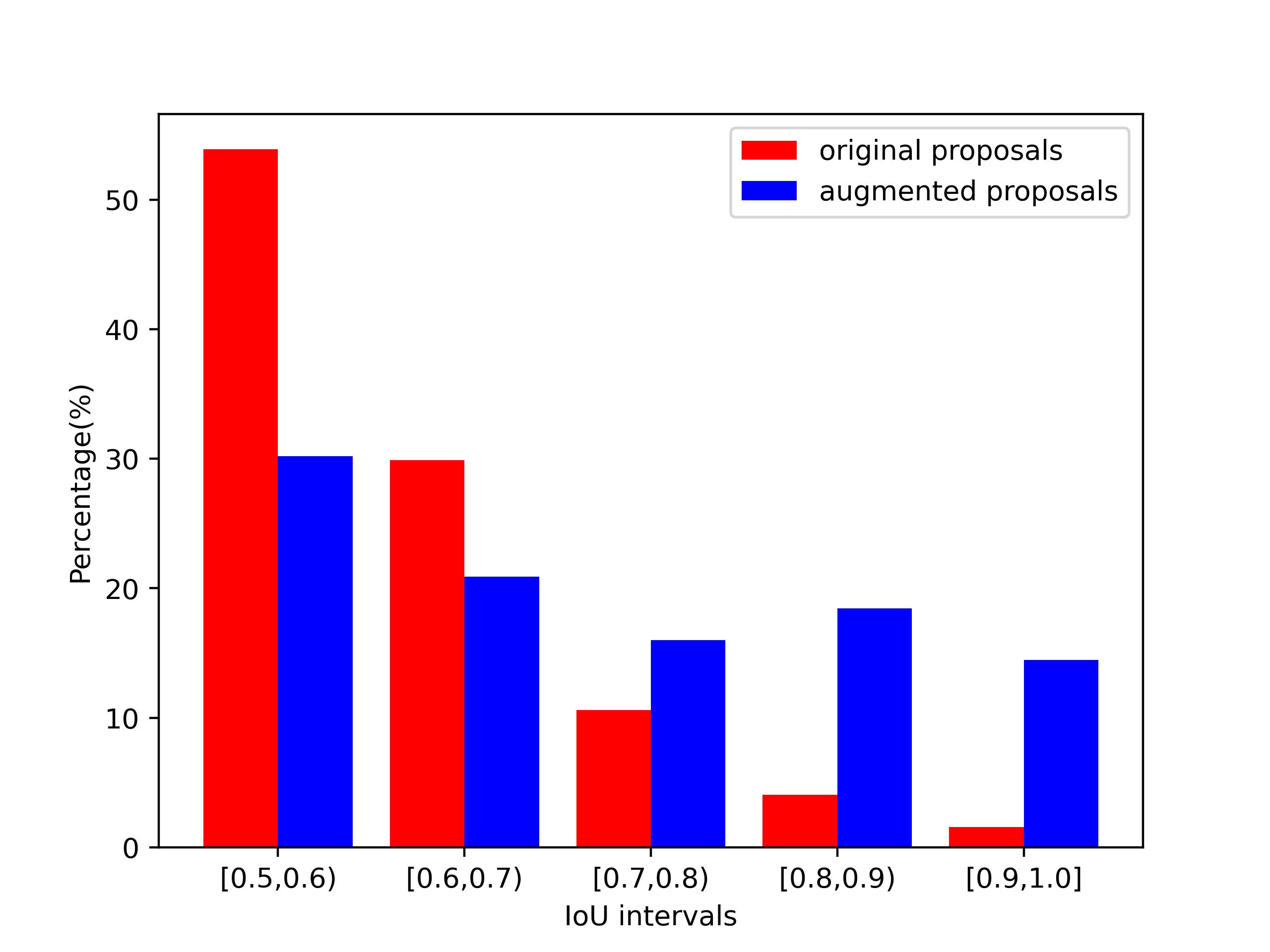}
    \caption{IoU distribution of original and augmented positive proposals.}
    \label{iou_distribution}
\end{figure}

Object detection is a fundamental and critical problem in computer vision, which needs to localize and recognize each object 
in images or videos. With the development of deep learning, current object detectors usually base on convolution neural 
networks, like FCOS~\cite{tian2019fcos}, ATSS~\cite{zhang2020bridging}, Fast\cite{girshick2015fast}/Faster\cite{ren2016faster} 
RCNN, Cascade R-CNN\cite{cai2018cascade}, DETR~\cite{carion2020end} and Sparse R-CNN~\cite{sun2020sparse}. They can be 
roughly divided into single-, two-, multi-stage and end-to-end methods based on their ways to generate detection results. 
Single-stage object detectors directly predict bounding boxes and corresponding scores for each location, while two- and 
multi-stage ones firstly generate proposals by Region Proposal Networks (RPN~\cite{ren2016faster}), then two-stage detectors 
use a RoI box head to refine the proposals and predict the category for each proposal while multi-stage detectors use multiple 
RoI box heads to do prediction. Single-, two- and multi-stage methods are dense detectors, so they need to apply non-maximum suppression 
(NMS) to filter redundant results, while end-to-end methods directly generate detection results without NMS. Typically, 
performance of multi-stage methods is better than two-stage and one-stage ones, since they apply coarse to fine strategy 
multi-times which is powerful to generate accurate bounding boxes and can precisely predict the category for each proposal.

For deep learning, if we regard the algorithm as the core of an object detector, then training samples are the power of it. 
Nearly all experiments on deep learning demonstrate that more training samples, better performance. Furthermore, diversity of 
training samples is also critical, which means that they need to be distributed evenly in a domain. For two-stage and multi-stage 
object detectors, the proposals generated by a RPN are the training samples of the RoI box heads, so, proposals' quality will 
significantly affect the RoI box head. There are two ways to improve proposals' quality. One is improving performance of RPN, 
for example, many detectors place more anchors with different scales and ratios in each location, which makes more anchors 
match the ground truth to get high recall rate. GA-RPN~\cite{wang2019region} firstly estimates anchors' 
location and shape, and then refines the estimated anchors to generate proposals. Cascade RPN~\cite{vu2019cascade} firstly 
refines the hand-designed anchors, and then aligns features of the refined anchors to process them again. These methods not only 
improve proposals' quality, but also make proposals' intersection over union(IoU) with ground truth more evenly distributed in 
0.5 to 1.0, which is critical to proposal diversity. The other is manually generating proposals, for example, 
IoU Net~\cite{jiang2018acquisition} manually transforms ground truth bounding boxes with a set of random parameters to obtain the 
proposals whose IoUs are evenly distributed in 0.5 to 1.0.

We hold the idea that the performance bottleneck of two-stage object detector is lacking of high quality proposals. As shown 
in Figure~\ref{iou_distribution}, few original positive proposals whose IoUs are greater than 0.8. However, placing more anchors in each 
location usually brings little improvements. While GA-RPN and Cascade RPN usually need about 150\% 
extra training costs, because they employ deformable operation\cite{dai2017deformable} to align the features for learned or refined 
anchors during their second step. Meanwhile, manually generated proposals do not match the distribution of 
proposals generated by the RPN, which may decrease robustness of detectors. 

For two-stage object detectors, the outputs 
of RoI heads naturally contain the characteristics of proposals, inspired by that, we develop a novel and simple method to augment 
proposals for RoI box head, which is named as APDI, augmenting proposals by the detector itself. APDI takes proposals generated by 
a RPN as coarse proposals, and then applies a RoI box head to refine them to obtain augmented ones. APDI does not add any parameters 
or hyper-parameters, but changes the proposals' generating process, and it only needs little extra training and inference costs. 
Extensive experiments on MS COCO dataset~\cite{lin2014microsoft} show that APDI can effectively improve performance of two-stage and multi-stage object 
detectors. Furthermore, with APDI, we can easily integrate the IoU head of IoU Net into a RoI box head, which merely adds subtle extra FLOPs 
since only an extra Linear layer is added. In this way, we can achieve at least 2.7 AP improvements on Faster R-CNN, and 1.1 AP increments for 
Cascade R-CNN. Furthermore, it can cooperate with GA-RPN and Cascade RPN and brings at least 4.0 AP improvements on Faster R-CNN. Our 
main contributions can be summarized as follows:

\begin{itemize}
    \item We propose a novel and simple method, APDI, which augments proposals by the detector itself and can significantly improve 
        their quality. It can not only be applied on two- and multi-stage object detectors, but also on instance segmentation methods.
        Furthermore, it can cooperate with GA-RPN and Cascade RPN.
    \item With APDI, we can easily integrate the IoU head into a RoI box head to predict an IoU score for each instance without 
        manually generated proposals, and obtain big improvements with subtle extra FLOPs.
    \item On MS COCO dataset, our method achieves at least 2.7 AP and 1.1 AP improvements on Faster R-CNN and Cascade R-CNN, 
        respectively, without any bells and whistles. 
\end{itemize}

\section{Related Work}

Generating proposals first, then refining proposals and predicting the category for each one has been a popular pipeline in object 
detection task. This kind of coarse to fine strategy usually brings surprising improvements. R-CNN~\cite{girshick2014rich}, 
Fast R-CNN~\cite{girshick2015fast} and Faster R-CNN~\cite{ren2016faster} all follow this design philosophy, but they also have some 
differences. R-CNN and Fast R-CNN employ Selective Search\cite{uijlings2013selective} to generate proposals, while Faster R-CNN 
designs a RPN which is based on deep learning to generate proposals and achieves joint training. 

RPN's performance greatly surpasses traditional proposal generating methods, like Selective Search and Edge Boxes~\cite{zitnick2014edge}, 
and it can get over 90\% recall rate when the IoU threshold is 0.5. However, most proposals' IoUs are distributed in 0.5 to 0.8, which means 
that the quality of the generated proposals is not good. GA-RPN separates the RPN's process into two steps. It firstly estimates the 
location and shape for each anchor, then it uses these learned anchors to match the ground truth. The following procedure is similar as 
RPN except it needs to align the feature for each learned anchor. GA-RPN greatly improves recall rate since its learned anchors can 
better match the ground truth than hand-designed anchors. Cascade RPN firstly refines the hand-designed anchors as the anchors of the 
next step, and then applies Adaptive Convolution to align the features to refine the anchors. Cascade RPN can further improve recall 
rate of proposals. 

Proposals of GA-RPN and Cascade RPN greatly improve quality of proposals, and they bring significant improvements on Faster R-CNN and
Cascade R-CNN. However, as Table~\ref{rpn_compare} shows, their AR drop fast with higher IoU thresholds, which means that they 
lack high quality proposals. Furthermore, they employ deformable operation to align features, which needs about 150\% extra training 
costs because they will encounter the writing lock during gradient computation.

Cascade R-CNN progressively refines proposals stage by stage, so proposals' quality will be gradually improved during this process since 
its next RoI head takes the outputs of its previous one as inputs. In this way, Cascade R-CNN obtains diverse training samples for RoI 
box heads, which is the key factor that it achieves significant improvements over Faster R-CNN. However, it needs about 30\% extra 
training costs since all three RoI box heads need to train, and if we ensemble collected proposals from different RoI box heads to train 
three heads, it will be a complex and time-consuming problem.

Recognizing detection quality by a model is an important factor, IoU Net proposes an IoU head to predict an IoU score for each detection 
result, and then uses this IoU score as the ranking keyword to guide the following NMS. In order to train its IoU head, it proposes 
a manual method which transforms the ground truth into proposals, and keeps their IoUs evenly distributed in 0.5 to 1.0. However, 
manually generated proposals may not match the distribution of proposals generated by RPN, so the manually generated ones could not 
directly replace the proposals during training the RoI box head. 

\begin{figure*}[t]
    \centering
    \includegraphics[width=0.66\linewidth]{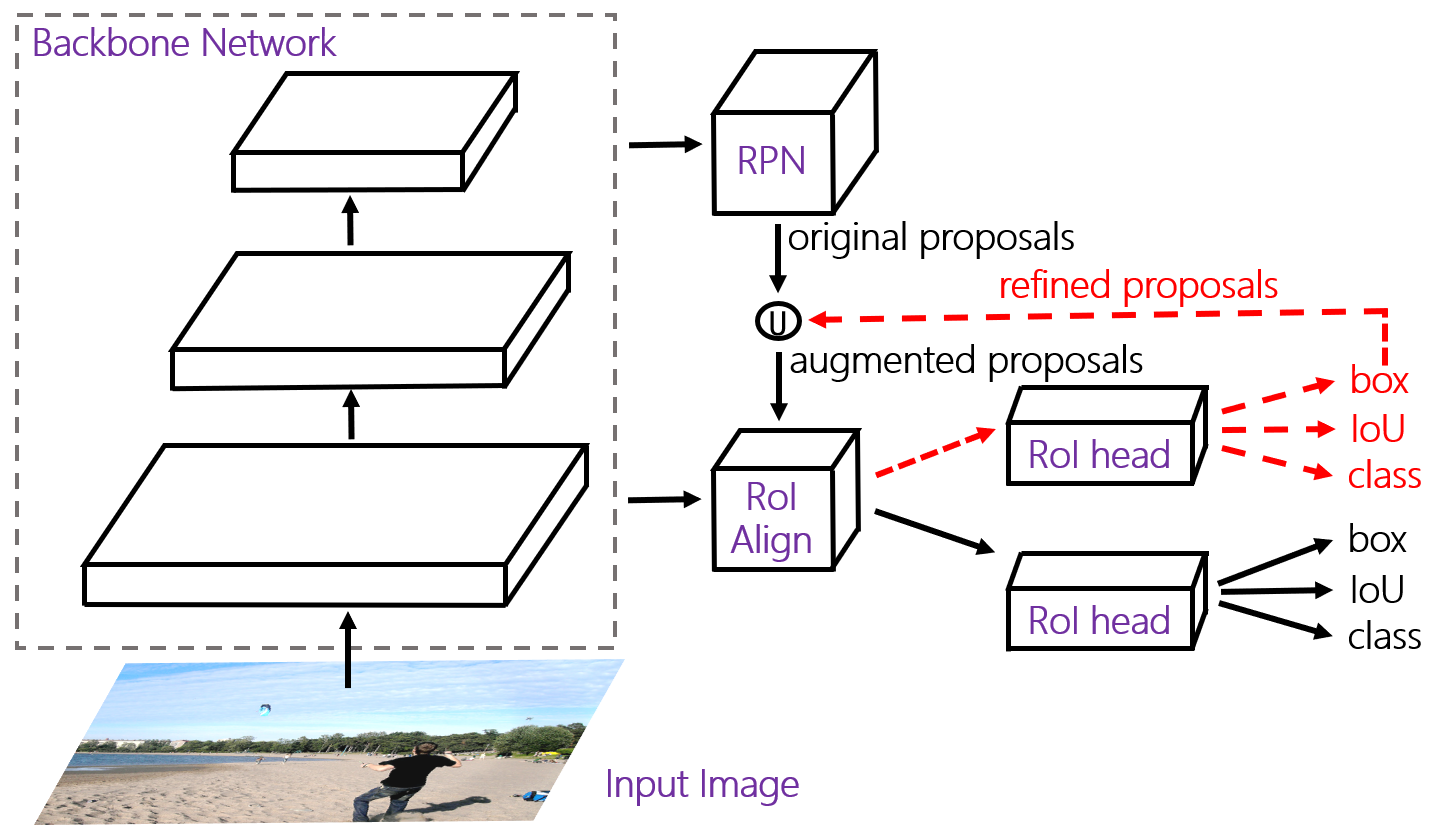}
    \caption{The pipeline of APDI, here the dotted red lines means no gradient will flow on the path.}
    \label{APDI}
\end{figure*}

\section{Method}

Proposals are critical for RoI box head training, so, how to collect enough higher quality proposals is an important factor. Unlike 
GA-RPN and Cascade RPN which design complicated pipeline for RPN to generate better proposals, APDI just runs the RoI box head with 
original proposals (generated by RPN) as inputs, and uses its outputs as refined proposals. Furthermore, APDI concatenate all the 
positive original proposals and the refined ones as augmented proposals. We use augmented proposals as input of the RoI box head in 
training and refined proposals in inference period. In order to align the training and inference process, we deploy box classification 
task on refined proposals, and box regression task on positive augmented proposals. 

APDI's pipeline is shown as Figure~\ref{APDI}, and its PyTorch-like 
Pseudocode is shown in Algorithm~\ref{alg_apdi}. In this section, we will firstly introduce our motivation, and then show more 
details about APDI, furthermore, we will integrate the IoU head of IoU Net into the RoI box head to predict the IoU scores to further
improve the performance, and save training and inference costs.

\subsection{Motivation}

\begin{table}
    \begin{center}
        \begin{tabular}{lccccc}
            \hline
            Method      & AR$_{50}$ & AR$_{60}$ & AR$_{70}$ & AR$_{80}$ & AR$_{90}$\\
            \hline
            RPN          & 92.7      & 88.9      & 78.7      & 46.1      & 8.1\\
            GA-RPN       & 94.8      & 91.8      & 84.8      & 65.1      & 27.6\\
            Cascade RPN  & 94.7      & 92.0      & 86.8      & 73.0      & 36.0\\
            RoI box head & 93.4      & 90.5      & 85.0      & 73.4      & 45.0\\
            \hline
        \end{tabular}
    \end{center}
    \caption{Comparisons on different region proposal networks.}
    \label{rpn_compare}
\end{table}

\begin{algorithm}[tb]
    \caption{Pseudocode of ADPI in a PyTorch-like style.}
    \label{alg_apdi}
    \begin{lstlisting}[gobble=4,basicstyle=\scriptsize\ttfamily]
    # ori_proposals: original proposals generated by RPN
    def augment_proposals(proposals)
        # detach the gradient
        with torch.no_grad():
            # refine original proposals by RoI box head
            ref_proposals = RoI_head(ori_proposals)
        # extract the positive samples from original proposals
        ious = compute_IoUs(ori_proposals)
        pos_proposals = ori_proposals[ious >= 0.5]
        # concatenate the positive and refined proposals
        aug_proposals = torch.cat([pos_proposals, ref_proposals])
        return aug_proposals
    \end{lstlisting}
\end{algorithm}

RPN's performance usually substantially affects the performance of RoI box head. We investigate the performance of RPN, GA-RPN and 
Cascade RPN. Here we employ COCO-style recall as the evaluation metrics, in the following, AR is the average of recall at different 
IoU thresholds, and AR$_x$ means recall at IoU threshold of $x$\%. The results are shown in Table~\ref{rpn_compare}, and we can 
find that what makes GA-RPN and Cascade RPN success is mainly because their proposals' quality is higher than RPN, in other words, 
their AR$_{70}$, AR$_{80}$ and AR$_{90}$ highly surpass those of RPN. However, if we follow the Faster R-CNN's pipeline 
to employ the RoI box head to refine the proposals, without any post-processing, we obtain comparable proposals ({\bf RoI box head} in 
Table~\ref{rpn_compare}) with GA-RPN and Cascade RPN. Therefore, if we directly employ RoI box head to augment the proposals before 
training, we may obtain comparable performance with complicated RPN, such as GA-RPN or Cascade RPN.

\begin{table*}
    \begin{center}
        \begin{tabular}{lllcccccc}
            \hline
            Method               & Backbone       & Schedule & AP & AP$_{50}$ & AP$_{75}$ & AP$_s$ & AP$_m$ & AP$_l$ \\
            \hline
            FCOS~\cite{tian2019fcos}                 & ResNet-101 FPN & - & 41.5 & 60.7 & 45.0 & 24.4 & 44.8 & 51.6\\               
            ATSS ~\cite{zhang2020bridging}           & ResNet-101 FPN & - & 43.6 & 62.1 & 47.4 & 26.1 & 47.0 & 53.6\\               
            FPN~\cite{lin2017feature}                  & ResNet-101 FPN & - & 36.2 & 59.1 & 39.0 & 18.2 & 39.0 & 48.2\\               
            Cascade R-CNN~\cite{cai2018cascade}        & ResNet-101 FPN & - & 42.8 & 62.1 & 46.3 & 23.7 & 45.5 & 55.2\\               
            IoU Net$\dagger$~\cite{jiang2018acquisition}     & ResNet-50 FPN  & - & 38.1 & 56.3 & -    & -    & -    & -   \\               
            Guided Anchoring~\cite{wang2019region}     & ResNet-50 FPN  & 1x & 39.8 & 59.2 & 43.5 & 21.8 & 42.6 & 50.7\\
            Cascade RPN~\cite{vu2019cascade}          & ResNet-50 FPN  & 1x & 40.6 & 58.9 & 44.5 & 22.0 & 42.8 & 52.6\\
            DETR$\dagger$~\cite{carion2020end}          & ResNet-50    & - & 42.0 & 62.4 & 44.2 & 20.5 & 45.8 & 61.1\\               
            Sparse R-CNN$\dagger$~\cite{sun2020sparse}  & ResNet-50    & 3x  & 42.3 & 61.2 & 45.7 & 26.7 & 44.6 & 57.6\\             
            \hline
            Faster R-CNN*        & ResNet-50 FPN  & 1x & 38.3 & 59.5 & 41.4 & 22.3 & 40.7 & 47.9\\      
            Cascade R-CNN*       & ResNet-50 FPN  & 1x & 41.7 & 59.7 & 45.2 & 24.1 & 44.0 & 52.6\\      
            Faster R-CNN*        & ResNet-50 FPN  & 3x & 40.4 & 61.4 & 43.9 & 23.6 & 42.8 & 50.3\\      
            Faster R-CNN*        & DCN-50 FPN     & 3x & 42.2 & 63.3 & 46.0 & 24.8 & 44.4 & 53.2\\      
            Faster R-CNN*        & ResNet-101 FPN & 3x & 42.4 & 63.1 & 46.1 & 25.0 & 45.0 & 53.4\\      
            \hline
            Faster R-CNN+APDI$^+$    & ResNet-50 FPN  & 1x & {\bf 42.0} & 60.4 & 45.4 & 24.4 & 44.2 & 53.1\\      
            Cascade R-CNN+APDI$^+$   & ResNet-50 FPN  & 1x & {\bf 42.8} & 60.2 & 46.1 & 24.9 & 44.8 & 54.3\\      
            Faster R-CNN+APDI$^+$    & ResNet-50 FPN  & 3x & {\bf 43.5} & 62.4 & 47.0 & 25.6 & 45.6 & 55.1\\      
            Faster R-CNN+APDI$^+$    & DCN-50 FPN     & 3x & {\bf 44.9} & 64.0 & 48.3 & 26.7 & 47.0 & 57.0\\      
            Faster R-CNN+APDI$^+$    & ResNet-101 FPN & 3x & {\bf 45.2} & 64.2 & 48.5 & 26.9 & 48.0 & 57.0\\      
            \hline
        \end{tabular}
    \end{center}
    \caption{Comparisons with state-of-the-art methods on COCO {\it test-dev}, the results with symbol '*' are our reimplemented results, and 
      the symbol '$\dagger$' means the results are reported on COCO {\it val-2017}. All settings 
      including '1x' and '3x' training schedules follow the detectron2 unless specially noted. '+ADPI$^+$' indicates 
      to apply APDI and Box IoU head on corresponding methods.}
    \label{comparison_sota}
\end{table*}



\subsection{Box IoU head}

\begin{figure}[t]
    \centering
    \subfigure[RoI box head + IoU head.]{\label{box_iou_head:a}\includegraphics[width=0.8\linewidth]{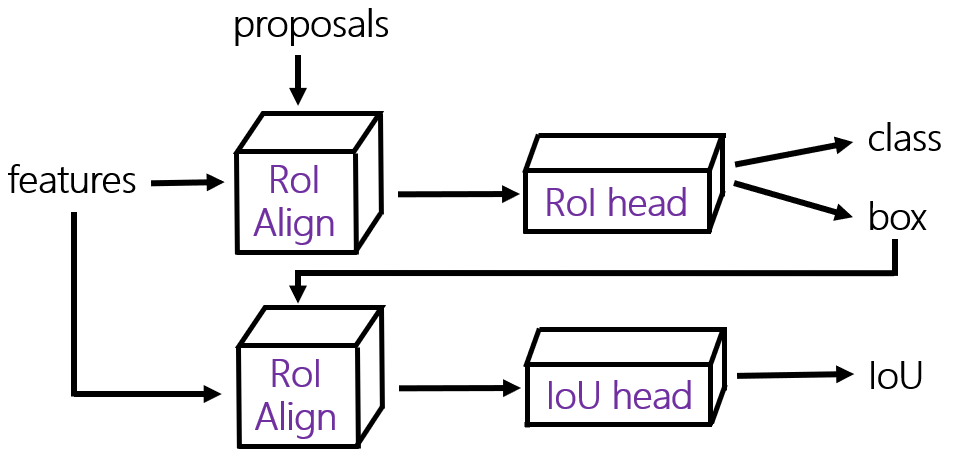}}
    \subfigure[Box IoU head.]{\label{box_iou_head:b}\includegraphics[width=0.8\linewidth]{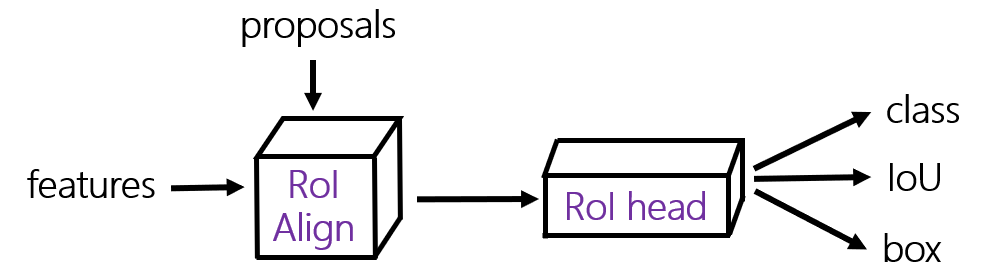}}
    \caption{{\bf RoI box head + IoU head} and {\bf Box IoU head}.}
    \label{box_iou_head}
\end{figure}

IoU Net demonstrates that IoU scores of detection results can guide the NMS and significantly improve the performance. However, it 
needs manually generated proposals to train its IoU head and takes about extra 15\% training and inference costs.

As shown in Figure~\ref{box_iou_head}, we integrate the IoU head into RoI box head, which is named as Box IoU head. We assume that 
proposals with higher IoU will obtain detection result with higher IoU, however, it is not true for original proposals, but is true 
for APDI. So, the targets of the IoU branch in Figure~\ref{box_iou_head:b} are IoUs of augmented proposals with ground truth. There 
are two reasons for this design. First, it is easily to compute the ground truth for IoU branch. Second, unlike RoI box head+IoU head, 
the quality of a proposal's detection result is not known until the proposal is sent into the Box IoU head, which is hard to optimize.  

During training, we use BCE as loss function for optimization of IoU scores, and the loss weight is simply set to 1.0, furthermore, 
all proposals whose IoU is greater or equal to 0.3 need to participate training.

During inference, the detection scores are product of box classification scores and IoU scores. The IoU scores are category-agnostic, 
which means that each detection has only one IoU score. Therefore, for each detection result, all detection scores for each category need to 
be calibrated by its IoU score.

\section{Experiments}

\subsection{Dataset and Evaluate Metrics}

All our experiments are implemented on the challenging Microsoft COCO 2017~\cite{lin2014microsoft} dataset. It consists of 118k images 
for training, {\it train-2017} and 5k images for validation, {\it val-2017}. We remove 1021 images with no usable annotations in 
{\it train-2017} following detectron2, thus 117k images are left. There are also 20k images without annotations for test in {\it test-dev}. 

We train all our models on {\it train-2017}, and report ablation studies on {\it val-2017} and final results on {\it test-dev}. All 
results are subjected to standard COCO-style Average Precision (AP) metrics which include AP (mean of AP over all IoU thresholds), 
AP$_{50}$, AP$_{75}$, AP$_s$, AP$_m$ and AP$_l$, where AP$_{50}$ and AP$_{75}$ are AP for IoU threshold of 50\% and 75\%, respectively, and 
AP$_s$, AP$_m$ and AP$_l$ are AP for small, medium, and large objects, respectively.

\subsection{Implementation Details}\label{details}

\begin{table*}
    \begin{center}
        \begin{tabular}{cc|cccccccccccc}
            \hline
            APDI & Box IoU head & AP & AP$_{50}$ & AP$_{75}$ & AP$_s$ & AP$_m$ & AP$_l$\\
            \hline
                       &            & 37.8 & 58.4 & 41.1 & 22.2 & 41.4 & 48.4\\
            \checkmark &            & 39.9 & 59.9 & 43.1 & 23.5 & 43.4 & 52.0\\
                       & \checkmark & 36.6 & 56.8 & 39.8 & 21.9 & 40.1 & 46.9\\
            \checkmark & \checkmark & 41.5 & 60.0 & 44.8 & 24.1 & 44.6 & 54.7\\
            \hline
        \end{tabular}
    \end{center}
    \caption{Comparison results for different components of this paper.}
    \label{ablation_study}
\end{table*}

All experiments are based on detectron2~\cite{wu2019detectron2}. And all our codes are deployed on a machine with 8 Tesla 
V100-SXM2-16GB GPUs and 32 Intel Xeon Platinum 8163 CPUs. Our software environment mainly include Ubuntu 18.04 LTS OS, CUDA 10.1, GCC 7.3, 
PyTorch~\cite{paszke2017automatic} 1.6.0 and Python 3.8.5. The other settings follow detectron2 unless specifically noted.

For Cascade R-CNN, we employ the first RoI box head to augment the proposals, and we also replace three RoI box heads with our proposed Box IoU head.
The original IoU thresholds of three RoI box heads of Cascade R-CNN are 0.5, 0.6 and 0.7, respectively. However, according to the IoUs' distribution 
of augmented proposals, we replace them by 0.5, 0.65 and 0.8, respectively. All other settings are same as APDI for Faster R-CNN.

Following a common practice~\cite{girshick2014rich}, all network backbones are pretrained on ImageNet1k classification set~\cite{5206848}, 
and their pre-trained models are publicly available.

All our models in Table~\ref{comparison_sota} are based on FPN~\cite{lin2017feature}. The first part of Table~\ref{comparison_sota} 
shows the results reported in original publications, the second part are our reimplemented results, and the last part are our methods.

\subsection{Results}

We apply APDI$^+$ (APDI+Box IoU head) on Faster R-CNN and Cascade R-CNN with ResNet-50~\cite{he2016deep}, ResNet-101~\cite{he2016deep} and 
DCN-50\cite{dai2017deformable}, and report our results on COCO {\it test-dev}. It is hard to compare different detectors completely 
fair due to different settings in training and testing, for example, batch-size, learning rate schedule and iterations, etc. Thus, we
reimplement them with various backbones and learning rate schedules, and other settings of them followed detectron2.

As shown in Table~\ref{comparison_sota}, with 1x training schedule, APDI$^+$ with ResNet-50 as backbone can achieve 42.0 AP without any 
bells and whistles, which is 3.7 AP higher than reimplemented Faster R-CNN with the same backbone. And Cascade R-CNN with APDI$^+$ achieves 
1.1 AP improvements. Furthermore, with 3x schedule, APDI$^+$ brings at least 2.7 AP improvements on Faster R-CNN with various backbones. 

\subsection{Ablation Study}

We report our ablation studies' results on COCO {\it val2017} dataset. Here, all base models are Faster R-CNN (ResNet-50 FPN) unless specially noted.

\subsubsection{APDI.}

From Table~\ref{ablation_study}, we find that APDI brings 2.1 AP improvements without any other tricks, and the AP$_{75}$ is 
largely improved, which means that the augmented proposals' quality has been significantly improved.

\subsubsection{Box IoU Head.}

From Table~\ref{ablation_study}, with Box IoU head only, we can find that the performance of the detector drops largely, which means that 
proposals' quality has no correlation with final detection results. However, it is interesting that if we combine APDI and Box IoU head 
together, then our proposed Box IoU head will bring extra 1.6 AP improvements. To explore underlying reasons, we visualize the distribution 
of original proposals and augmented positive proposals in Figure~\ref{iou_distribution}. We find that the IoUs of the augmented proposals are 
more evenly distributed in 0.5 to 1.0, while IoUs of the original proposals are mainly distributed in 0.5 to 0.8. Thus, directly applying these 
unbalanced proposals to train the IoU branch will lead to poor performance according to IoU Net.

\subsubsection{Cascade R-CNN.}

\begin{table}
    \begin{center}
        \begin{tabular}{ccccccc}
            \hline
            APDI & change & Box IoU head & AP & AP$_{50}$ & AP$_{75}$\\
            \hline
                       &            &            & 41.6 & 59.5 & 45.1\\
            \checkmark &            &            & 41.8 & 59.7 & 45.2\\
            \checkmark & \checkmark &            & 42.1 & 59.2 & 45.5\\
            \checkmark & \checkmark & \checkmark & 42.5 & 59.7 & 46.1\\
            \hline
        \end{tabular}
    \end{center}
    \caption{Comparison for different strategies on Cascade R-CNN.}
    \label{apdi_cascade}
\end{table}

For Cascade R-CNN, as mentioned in Section~\ref{details}, we change the IoU thresholds of its three RoI box heads ({\bf change} in Table~\ref{apdi_cascade}), 
and replace its RoI box heads with three Box IoU heads. In this section, we conduct ablation study for these operations, and report the results in 
Table~\ref{apdi_cascade}.

From Table~\ref{apdi_cascade}, we find that directly applying APDI on Cascade R-CNN brings 0.2 AP improvements, and 0.3 AP improvements could
be obtained if we change the IoU thresholds. So, Cascade R-CNN's original IoU thresholds are inappropriate for our high quality proposals, 
while higher IoU thresholds for the second and third head are more reasonable since the number of high quality proposals is increased. 
Furthermore, with Box IoU head, we obtain extra 0.4 AP improvements, which means more fine quality scores are also useful for Cascade R-CNN.



\subsubsection{Training and Inference Costs.}

\begin{table*}
    \begin{center}
        \begin{tabular}{lcccccccc}
            \hline
            Method & training time(ms/iter) & inference time(ms/img) & AP & AP$_{50}$ & AP$_{75}$ & AP$_s$ & AP$_m$ & AP$_l$\\
            \hline
            RPN         & 189.0 & 40.5 & 37.8 & 58.4 & 41.1 & 22.2 & 41.4 & 48.4\\
            GA-RPN      & 448.1 & 69.3 & 39.6 & 58.7 & 43.4 & 21.2 & 43.0 & 52.7\\
            Cascade RPN & 436.4 & 67.6 & 40.4 & 59.1 & 43.9 & 22.9 & 43.5 & 53.0\\
            \hline
            RPN+APDI    & 204.3 & 45.3 & 39.9 & 59.9 & 43.1 & 23.5 & 43.4 & 52.0\\
            \hline
        \end{tabular}
    \end{center} 
    \caption{Comparison of different RPNs.}
    \label{speed}
\end{table*}

GA-RPN and Cascade RPN are originally implemented in mmdetection2~\cite{mmdetection}, while all our codes implemented in detectron2, 
so there may exist some performance gap. To get fair comparison, here we directly call their APIs in mmdetection2 to reimplement them.
The performance of our reproduced results is slightly worse than the results in original publications. So we 
copy the results reported in mmdetection2, but test the training and inference time by our reproduced ones. We report the training 
and inference costs in Table~\ref{speed}.

We find that APDI is only slightly slower than standard RPN since APDI is a gradient free method, while GA-RPN and Cascade RPN's 
training time is more than twice because of writing lock. And for inference time, APDI is slightly slower than RPN by 12\%, while GA-RPN and 
Cascade RPN need extra 76\% and 71\% inference time, respectively. Furthermore, APDI achieves comparable AP with them, so, APDI is a 
cost-efficient method.

\section{Discussion}

In this section, we will deeply discuss differences with iterative bounding box regression (IBBR) and the cooperation with improved 
RPN, like GA-RPN and Cascade RPN. Furthermore, we also try to apply APDI on instance segmentation tasks. All following models are 
trained on COCO {\it train-2017} and the results are reported on COCO {\it val-2017} dataset. In this section, all base models are 
Faster R-CNN (ResNet-50 FPN) unless specially mentioned.

\subsection{IBBR}

\begin{table}
    \begin{center}
        \begin{tabular}{lcccccc}
            \hline
            Method & AP & AP$_{50}$ & AP$_{75}$ & AP$_s$ & AP$_m$ & AP$_l$\\
            \hline
            None   & 37.8 & 58.4 & 41.1 & 22.2 & 41.4 & 48.4\\
            IBBR   & 38.2 & 58.5 & 41.6 & 21.9 & 41.2 & 50.1\\
            APDI   & 39.9 & 59.9 & 43.1 & 23.5 & 43.4 & 52.0\\
            \hline
        \end{tabular}
    \end{center}
    \caption{Comparison for different strategies on Faster R-CNN.}
    \label{iterative}
\end{table}

IBBR means appling RoI box head several times to refine the bounding boxes during inference, which is similar to APDI in inference. 
However, the key difference is that APDI works on both training and inference period, but IBBR only works on inference. We apply 
bounding box regression 2 times for IBBR, and the results are shown in Table~\ref{iterative}.

From Table~\ref{iterative}, we can conclude that IBBR only brings little gains on Faster R-CNN, and it is beneficial for large instances, but harmful for small 
ones. Furthermore, experiments in IoU Net~\cite{jiang2018acquisition} show that increasing number of iterations will degrade performance of IBBR.  
However, APDI surpasses Faster R-CNN on all metrics, mainly because of the well-trained RoI box head by the augmented proposals.

\subsection{Cooperation with Improved RPN}

APDI tries to augment the proposals generated by RPN, furthermore, we find that APDI can also cooperate well with GA-RPN and Cascade RPN. 
We directly replace the RPN by them, and keep all other settings unchanged. All results are reported in 
Table~\ref{cooperate_with_ga_cascade_rpn}, and we can draw a conclusion that GA-RPN or Cascade RPN and APDI are not competitive, but 
complementary. Moreover, the gains of GA-RPN and Cascade RPN mainly come from AP$_{75}$, which means that better proposals 
usually lead to more precise detection results.

Here, we also find that Box IoU head brings 1.6 AP improvements on RPN, while 1.1 AP and 1.0 AP improvements on GA-RPN and Cascade RPN,
respectively. It is possibly because there are gaps between IoU distribution of GA-RPN or Cascade RPN and RPN, which bring a large bias 
for predicting IoU scores.

\begin{table}
    \begin{center}
        \begin{tabular}{lcccc}
            \hline
            Method      & Box IoU head & AP & AP$_{50}$ & AP$_{75}$\\
            \hline
            RPN         &            & 39.9 & 59.9 & 43.1\\
            GA-RPN      &            & 40.7 & 59.3 & 44.3\\
            Cascade RPN &            & 41.3 & 59.7 & 44.8\\
            RPN         & \checkmark & 41.5 & 60.4 & 44.9\\
            GA-RPN      & \checkmark & 41.8 & 60.1 & 45.2\\
            Cascade RPN & \checkmark & 42.3 & 60.2 & 45.7\\
            \hline
        \end{tabular}
    \end{center}
    \caption{Comparison for different RPNs with APDI.}
    \label{cooperate_with_ga_cascade_rpn}
\end{table}

\subsection{Instance Segmentation with APDI}


Instance segmentation needs to predict a mask for each instance, and we employ Mask R-CNN\cite{he2017mask} as our base model since the mask 
head's pipeline is similar to RoI box head. We use the positive refined proposals as inputs of the mask head to save training costs, and 
other settings are same as Faster R-CNN with APDI and Box IoU head. We also conduct our experiments on Cascade Mask R-CNN.

As shown in Table~\ref{segm}, for instance segmentation, APDI obtains 1.8 AP and 0.8 AP improvements on Mask R-CNN and Cascade Mask R-CNN, 
respectively. It shows that APDI is also powerful for instance segmentation tasks.

\begin{table}
    \begin{center}
        \begin{tabular}{lcccc}
            \hline
            Method      & APDI & AP$^{bbox}$ & AP$^{mask}$\\
            \hline
            Mask R-CNN         &            & 38.6 & 35.2\\
            Mask R-CNN         & \checkmark & 42.0 & 37.0\\
            Cascade Mask R-CNN &            & 42.2 & 36.6\\
            Cascade Mask R-CNN & \checkmark & 43.3 & 37.4\\
            \hline
        \end{tabular}
    \end{center}
    \caption{Comparison on instance segmentation tasks with different methods.}
    \label{segm}
\end{table}

\section{Conclusion}

In this paper, we propose a novel and simple training method named APDI, which can significantly improve performance with little 
extra FLOPs. Furthermore, with the help of APDI, we also find that we can integrate IoU head into RoI Box head, which is named as Box IoU 
head. It is concise and brings big improvements. We hope that the findings could help researchers design more powerful object detector. 
In our future work, we will deeply explore more advanced cooperation between APDI and improved RPN methods.

\bibliographystyle{named}
\bibliography{ijcai21}

\begin{thebibliography}{}

\bibitem[\protect\citeauthoryear{Cai and Vasconcelos}{2018}]{cai2018cascade}
Zhaowei Cai and Nuno Vasconcelos.
\newblock Cascade r-cnn: Delving into high quality object detection.
\newblock In {\em Proceedings of the IEEE conference on computer vision and
  pattern recognition}, pages 6154--6162, 2018.

\bibitem[\protect\citeauthoryear{Carion \bgroup \em et al.\egroup
  }{2020}]{carion2020end}
Nicolas Carion, Francisco Massa, Gabriel Synnaeve, Nicolas Usunier, Alexander
  Kirillov, and Sergey Zagoruyko.
\newblock End-to-end object detection with transformers.
\newblock {\em arXiv preprint arXiv:2005.12872}, 2020.

\bibitem[\protect\citeauthoryear{Chen \bgroup \em et al.\egroup
  }{2019}]{mmdetection}
Kai Chen, Jiaqi Wang, Jiangmiao Pang, Yuhang Cao, Yu~Xiong, Xiaoxiao Li,
  Shuyang Sun, Wansen Feng, Ziwei Liu, Jiarui Xu, Zheng Zhang, Dazhi Cheng,
  Chenchen Zhu, Tianheng Cheng, Qijie Zhao, Buyu Li, Xin Lu, Rui Zhu, Yue Wu,
  Jifeng Dai, Jingdong Wang, Jianping Shi, Wanli Ouyang, Chen~Change Loy, and
  Dahua Lin.
\newblock {MMDetection}: Open mmlab detection toolbox and benchmark.
\newblock {\em arXiv preprint arXiv:1906.07155}, 2019.

\bibitem[\protect\citeauthoryear{Dai \bgroup \em et al.\egroup
  }{2017}]{dai2017deformable}
Jifeng Dai, Haozhi Qi, Yuwen Xiong, Yi~Li, Guodong Zhang, Han Hu, and Yichen
  Wei.
\newblock Deformable convolutional networks.
\newblock In {\em Proceedings of the IEEE international conference on computer
  vision}, pages 764--773, 2017.

\bibitem[\protect\citeauthoryear{{Deng} \bgroup \em et al.\egroup
  }{2009}]{5206848}
J.~{Deng}, W.~{Dong}, R.~{Socher}, L.~{Li}, {Kai Li}, and {Li Fei-Fei}.
\newblock Imagenet: A large-scale hierarchical image database.
\newblock In {\em 2009 IEEE Conference on Computer Vision and Pattern
  Recognition}, pages 248--255, 2009.

\bibitem[\protect\citeauthoryear{Girshick \bgroup \em et al.\egroup
  }{2014}]{girshick2014rich}
Ross Girshick, Jeff Donahue, Trevor Darrell, and Jitendra Malik.
\newblock Rich feature hierarchies for accurate object detection and semantic
  segmentation.
\newblock In {\em Proceedings of the IEEE conference on computer vision and
  pattern recognition}, pages 580--587, 2014.

\bibitem[\protect\citeauthoryear{Girshick}{2015}]{girshick2015fast}
Ross Girshick.
\newblock Fast r-cnn.
\newblock In {\em Proceedings of the IEEE international conference on computer
  vision}, pages 1440--1448, 2015.

\bibitem[\protect\citeauthoryear{He \bgroup \em et al.\egroup
  }{2016}]{he2016deep}
Kaiming He, Xiangyu Zhang, Shaoqing Ren, and Jian Sun.
\newblock Deep residual learning for image recognition.
\newblock In {\em Proceedings of the IEEE conference on computer vision and
  pattern recognition}, pages 770--778, 2016.

\bibitem[\protect\citeauthoryear{He \bgroup \em et al.\egroup
  }{2017}]{he2017mask}
Kaiming He, Georgia Gkioxari, Piotr Doll{\'a}r, and Ross Girshick.
\newblock Mask r-cnn.
\newblock In {\em Proceedings of the IEEE international conference on computer
  vision}, pages 2961--2969, 2017.

\bibitem[\protect\citeauthoryear{Jiang \bgroup \em et al.\egroup
  }{2018}]{jiang2018acquisition}
Borui Jiang, Ruixuan Luo, Jiayuan Mao, Tete Xiao, and Yuning Jiang.
\newblock Acquisition of localization confidence for accurate object detection.
\newblock In {\em Proceedings of the European Conference on Computer Vision
  (ECCV)}, pages 784--799, 2018.

\bibitem[\protect\citeauthoryear{Lin \bgroup \em et al.\egroup
  }{2014}]{lin2014microsoft}
Tsung-Yi Lin, Michael Maire, Serge Belongie, James Hays, Pietro Perona, Deva
  Ramanan, Piotr Doll{\'a}r, and C~Lawrence Zitnick.
\newblock Microsoft coco: Common objects in context.
\newblock In {\em European conference on computer vision}, pages 740--755.
  Springer, 2014.

\bibitem[\protect\citeauthoryear{Lin \bgroup \em et al.\egroup
  }{2017}]{lin2017feature}
Tsung-Yi Lin, Piotr Doll{\'a}r, Ross Girshick, Kaiming He, Bharath Hariharan,
  and Serge Belongie.
\newblock Feature pyramid networks for object detection.
\newblock In {\em Proceedings of the IEEE conference on computer vision and
  pattern recognition}, pages 2117--2125, 2017.

\bibitem[\protect\citeauthoryear{Paszke \bgroup \em et al.\egroup
  }{2017}]{paszke2017automatic}
Adam Paszke, Sam Gross, Soumith Chintala, Gregory Chanan, Edward Yang, Zachary
  DeVito, Zeming Lin, Alban Desmaison, Luca Antiga, and Adam Lerer.
\newblock Automatic differentiation in pytorch.
\newblock 2017.

\bibitem[\protect\citeauthoryear{Ren \bgroup \em et al.\egroup
  }{2016}]{ren2016faster}
Shaoqing Ren, Kaiming He, Ross Girshick, and Jian Sun.
\newblock Faster r-cnn: Towards real-time object detection with region proposal
  networks.
\newblock {\em IEEE transactions on pattern analysis and machine intelligence},
  39(6):1137--1149, 2016.

\bibitem[\protect\citeauthoryear{Sun \bgroup \em et al.\egroup
  }{2020}]{sun2020sparse}
Peize Sun, Rufeng Zhang, Yi~Jiang, Tao Kong, Chenfeng Xu, Wei Zhan, Masayoshi
  Tomizuka, Lei Li, Zehuan Yuan, Changhu Wang, et~al.
\newblock Sparse r-cnn: End-to-end object detection with learnable proposals.
\newblock {\em arXiv preprint arXiv:2011.12450}, 2020.

\bibitem[\protect\citeauthoryear{Tian \bgroup \em et al.\egroup
  }{2019}]{tian2019fcos}
Zhi Tian, Chunhua Shen, Hao Chen, and Tong He.
\newblock Fcos: Fully convolutional one-stage object detection.
\newblock In {\em Proceedings of the IEEE international conference on computer
  vision}, pages 9627--9636, 2019.

\bibitem[\protect\citeauthoryear{Uijlings \bgroup \em et al.\egroup
  }{2013}]{uijlings2013selective}
Jasper~RR Uijlings, Koen~EA Van De~Sande, Theo Gevers, and Arnold~WM Smeulders.
\newblock Selective search for object recognition.
\newblock {\em International journal of computer vision}, 104(2):154--171,
  2013.

\bibitem[\protect\citeauthoryear{Vu \bgroup \em et al.\egroup
  }{2019}]{vu2019cascade}
Thang Vu, Hyunjun Jang, Trung~X Pham, and Chang Yoo.
\newblock Cascade rpn: Delving into high-quality region proposal network with
  adaptive convolution.
\newblock In {\em Advances in Neural Information Processing Systems}, pages
  1432--1442, 2019.

\bibitem[\protect\citeauthoryear{Wang \bgroup \em et al.\egroup
  }{2019}]{wang2019region}
Jiaqi Wang, Kai Chen, Shuo Yang, Chen~Change Loy, and Dahua Lin.
\newblock Region proposal by guided anchoring.
\newblock In {\em Proceedings of the IEEE Conference on Computer Vision and
  Pattern Recognition}, pages 2965--2974, 2019.

\bibitem[\protect\citeauthoryear{Wu \bgroup \em et al.\egroup
  }{2019}]{wu2019detectron2}
Yuxin Wu, Alexander Kirillov, Francisco Massa, Wan-Yen Lo, and Ross Girshick.
\newblock Detectron2.
\newblock \url{https://github.com/facebookresearch/detectron2}, 2019.

\bibitem[\protect\citeauthoryear{Zhang \bgroup \em et al.\egroup
  }{2020}]{zhang2020bridging}
Shifeng Zhang, Cheng Chi, Yongqiang Yao, Zhen Lei, and Stan~Z Li.
\newblock Bridging the gap between anchor-based and anchor-free detection via
  adaptive training sample selection.
\newblock In {\em Proceedings of the IEEE/CVF Conference on Computer Vision and
  Pattern Recognition}, pages 9759--9768, 2020.

\bibitem[\protect\citeauthoryear{Zitnick and
  Doll{\'a}r}{2014}]{zitnick2014edge}
C~Lawrence Zitnick and Piotr Doll{\'a}r.
\newblock Edge boxes: Locating object proposals from edges.
\newblock In {\em European conference on computer vision}, pages 391--405.
  Springer, 2014.

\end{thebibliography}

\end{document}